\documentclass{article}
\usepackage{spconf,amsmath,amssymb,graphicx,booktabs,cite}
\usepackage{enumitem}
\usepackage{microtype}
\usepackage{balance}
\usepackage{url}

\title{On the Robustness--Resolution Tradeoff in Temporal Quantization of Event Streams}
\name{Sayeed Shafayet Chowdhury$^1$, Ruhi Sharmin$^2$, Syed Ishtiaque Ahmed$^3$}
\address{$^1$Department of Computer Science, Luddy School of Informatics, Computing, and Engineering\\
Indiana University Indianapolis, Indianapolis, IN, USA\\
$^2$The Data Mine, Department of Computer Science, Purdue University, Indianapolis, IN, USA\\
$^3$Department of Computer Science, University of Toronto, Toronto, Ontario, Canada}

\begin{document}
\ninept
\maketitle

\begin{abstract}
Event pipelines often discretize asynchronous timestamps before learning. This step looks harmless, but its stability depends directly on temporal resolution. We study this dependence at the representation level. We first show that hard temporal binning is discontinuous: an arbitrarily small timestamp shift near a boundary can move unit event mass between bins. We then define a class of nonnegative, mass-preserving, resolution-faithful continuous encoders and prove that every encoder in this class has global $\ell_1$ sensitivity at least $2/\Delta$, where $\Delta$ denotes bin width. Linear two-bin interpolation attains this limit. Local support and first-moment preservation also make it unique. Experiments on SHD, N-MNIST, and DVS128 Gesture support the analysis. Across uniform timestamp budgets, linear interpolation lowers mean representation drift by 47--72\% while keeping clean accuracy nearly unchanged. On DVS Gesture, it produces zero prediction flips across all tested budgets and three seeds. On SHD, measured drift follows $1/\Delta$ with $R^2=0.992$.
\end{abstract}

\begin{keywords}
event streams, temporal quantization, neuromorphic computing, temporal stability, event representation
\end{keywords}

\section{Introduction}
Event sensors report sparse changes with precise timestamps rather than dense frames~\cite{lichtsteiner2008,gallego2022,ren2026survey}. Most learning pipelines still convert these asynchronous events to regular tensors before inference~\cite{ren2026survey,ye2025}. They sum events into frames, place them in voxel grids, or build learned event surfaces~\cite{hay2025,liang2025}. This conversion makes event streams easy to process with standard networks, but it also changes the time axis. A timestamp that crosses one bin boundary can alter a tensor even when the physical time shift stays tiny.

We study that preprocessing step directly, and ask a simple question: \emph{how much timestamp sensitivity must a temporally resolved encoder incur?} We separate this question from architecture design and task-specific attack optimization. We first analyze hard binning, and subsequently study continuous temporal encoders that preserve event mass and represent anchors spaced by $\Delta$ exactly. This setup exposes a resolution--stability tradeoff that depends only on the encoder.
Our main contributions are:
\begin{itemize}[leftmargin=1.25em,itemsep=0pt,topsep=2pt]
    \item We show that hard temporal binning has no finite global Lipschitz constant with respect to timestamp shifts. We also derive a sample-specific bound that counts only events close enough to cross a boundary.
    \item We prove that every nonnegative, mass-preserving continuous encoder with exact anchors spaced by $\Delta$ has global $\ell_1$ sensitivity at least $2/\Delta$. Linear two-bin interpolation reaches this lower bound.
    \item We show that local support and first-moment preservation uniquely recover the same linear encoder. We then derive a stream-level perturbation bound and a classifier-margin corollary.
    \item We test the analysis on spike-based speech, saccadic event vision, and real dynamic event vision. Linear interpolation cuts representation drift by 47--72\% across SHD~\cite{cramer2022}, N-MNIST~\cite{orchard2015}, and DVS128 Gesture~\cite{amir2017}, with little clean-accuracy change.
\end{itemize}

\section{Related Work}
\textbf{Event representations.}
Event pipelines convert asynchronous streams into event frames, histograms, time surfaces, voxel grids, or learned surfaces~\cite{gallego2022,ren2026survey,hay2025,araghi2025}. These choices preserve time in different ways. Time surfaces retain recency, while frame and histogram methods discard within-bin ordering. Sironi \emph{et al.} aggregate local time surfaces into HATS descriptors~\cite{sironi2018}. Zhu \emph{et al.} retain temporal order with a discretized event volume~\cite{zhu2019}. Gehrig \emph{et al.} place events on regular grids through differentiable interpolation kernels~\cite{gehrig2019}, and Cannici \emph{et al.} learn recurrent event surfaces~\cite{cannici2020}. Recent work also studies continuous multi-timescale B-spline encodings~\cite{lundell2026}. These methods mainly compare task accuracy, efficiency, or learnability~\cite{araghi2025,hay2025,liang2025}. They do not characterize the minimum timestamp sensitivity imposed by a requested temporal resolution.

\textbf{Timing perturbations and our position.}
Lee and Myung shift timestamps and add adversarial events to fool event classifiers~\cite{lee2022}. Yu \emph{et al.} study timing-only spike retiming that preserves spike counts and amplitudes in event-driven SNNs~\cite{yu2026}. These studies optimize attacks and measure network failure~\cite{lee2022,yu2026,lin2025}. We instead analyze the temporal map before the network, so our bounds do not depend on a classifier, loss, or attack direction. Linear interpolation itself is established in voxel-style event representations~\cite{gehrig2019,zhu2019}; our contribution is its stability characterization. We connect hard-bin discontinuity, a boundary-local hard-bin bound, a $2/\Delta$ lower bound for mass-preserving resolution-faithful continuous encoders, and minimax attainability, then test the same prediction across three event geometries and two noise laws.

\section{Temporal Encoding and Stability}
\subsection{Setup}
We write an event stream as $\mathcal{E}=\{(m_i,t_i,a_i)\}_{i=1}^N$. Here $m_i$ stores the event mark, such as a cochlear channel or a pixel-polarity index, $t_i\in\mathbb{R}$ stores time, and $a_i$ stores event amplitude. We study additive temporal encoders
\begin{equation}
R_\psi(\mathcal{E})=\sum_{i=1}^{N}a_i\,e_{m_i}\otimes\psi(t_i),
\label{eq:repr}
\end{equation}
where $\psi(t)\in\mathbb{R}^{B}$ maps one timestamp to $B$ temporal channels. We measure representation change with $\ell_1$ distance. In experiments, we normalize this distance by $M(\mathcal{E})=\sum_i|a_i|$.

\subsection{Hard Binning Creates a Jump}
Hard quantization with bin width $\Delta$ uses $\psi_{\rm H}(t)=e_{\lfloor t/\Delta\rfloor}$.

\textbf{Proposition 1 (discontinuity).} For every $\eta>0$, there exist $t,t'$ with $|t-t'|<\eta$ but $\|\psi_{\rm H}(t)-\psi_{\rm H}(t')\|_1=2$.

\textit{Proof.} Choose $t=k\Delta-\eta/4$ and $t'=k\Delta+\eta/4$. The timestamps fall on opposite sides of the same boundary, so hard binning returns two different one-hot vectors. Shrinking $\eta$ leaves the output jump at $2$. Hence no finite global Lipschitz constant can bound $\psi_{\rm H}$. $\square$

\begin{figure}[t]
\centering
\includegraphics[width=0.92\linewidth]{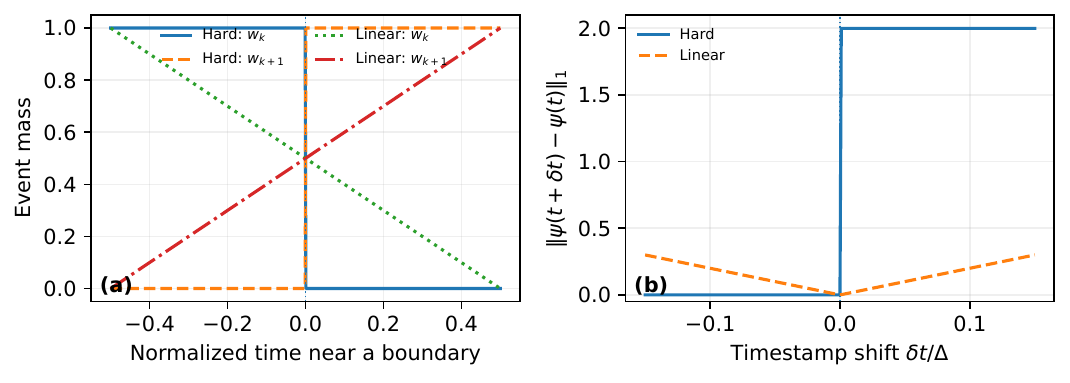}
\caption{Encoding near one temporal boundary. Hard binning switches all event mass at the boundary. Linear interpolation transfers mass continuously.}
\label{fig:encoding}
\vspace{-1.5mm}
\end{figure}

Figure~\ref{fig:encoding} shows the same effect. Hard binning creates a finite jump at the boundary. Linear interpolation changes continuously.

\subsection{A Boundary-Local Bound for Hard Bins}
Define the distance to the closest hard-bin boundary as
\begin{equation}
d_\Delta(t)=\min_{k\in\mathbb{Z}}|t-k\Delta|,
\end{equation}
and boundary mass as
\begin{equation}
B_\epsilon(\mathcal{E})=\sum_i |a_i|\,\mathbf{1}\!\left[d_\Delta(t_i)\le\epsilon\right].
\end{equation}
\textbf{Proposition 2 (boundary-local bound).} If $|\delta_i|\le\epsilon<\Delta/2$ for every event, then
\begin{equation}
\|R_{\rm H}(\mathcal{E}_\delta)-R_{\rm H}(\mathcal{E})\|_1\le 2B_\epsilon(\mathcal{E}).
\label{eq:hardlocal}
\end{equation}
\textit{Proof.} An event can change bins only when it starts within $\epsilon$ of a boundary. Each crossing moves mass $|a_i|$ between two one-hot coordinates and contributes at most $2|a_i|$ in $\ell_1$. Summing the eligible events proves~\eqref{eq:hardlocal}. $\square$

This local result does not remove Proposition~1. For any arbitrarily small budget, an event can still lie arbitrarily close to a boundary and incur the full jump.

\subsection{A Resolution--Stability Lower Bound}
Let $\mathcal{A}_\Delta$ contain continuous temporal maps $\psi$ that satisfy
\begin{equation}
\psi(t)\ge 0,\qquad \|\psi(t)\|_1=1,\qquad \psi(k\Delta)=e_k.
\label{eq:class}
\end{equation}
The first two conditions preserve nonnegative event mass. The last condition makes the encoder exact at temporal anchors. Define
\begin{equation}
L_\psi=\sup_{s\ne t}\frac{\|\psi(t)-\psi(s)\|_1}{|t-s|}.
\end{equation}

\textbf{Theorem 1 (resolution--stability bound).} Every $\psi\in\mathcal{A}_\Delta$ with finite $L_\psi$ satisfies
\begin{equation}
L_\psi\ge \frac{2}{\Delta}.
\label{eq:lower}
\end{equation}
\textit{Proof.} Adjacent anchors give $\psi(k\Delta)=e_k$ and $\psi((k+1)\Delta)=e_{k+1}$. Their $\ell_1$ distance equals $2$, while their temporal separation equals $\Delta$. The definition of $L_\psi$ gives~\eqref{eq:lower}. $\square$

For $t\in[k\Delta,(k+1)\Delta]$, let $\alpha=(t-k\Delta)/\Delta$. Linear interpolation uses
\begin{equation}
\psi_{\rm L}(t)=(1-\alpha)e_k+\alpha e_{k+1}.
\label{eq:linear}
\end{equation}
Inside each interval, $\|d\psi_{\rm L}/dt\|_1=2/\Delta$. Continuity across anchors therefore gives
\begin{equation}
\|\psi_{\rm L}(t)-\psi_{\rm L}(s)\|_1\le \frac{2}{\Delta}|t-s|.
\end{equation}
Together with Theorem~1,
\begin{equation}
\inf_{\psi\in\mathcal{A}_\Delta}L_\psi=L_{\psi_{\rm L}}=\frac{2}{\Delta}.
\label{eq:minimax}
\end{equation}
\textbf{Corollary 1 (minimax sensitivity).} Linear interpolation minimizes worst-case global $\ell_1$ timestamp sensitivity over $\mathcal{A}_\Delta$. This statement does not claim that it optimizes task accuracy or average noise response.

\textbf{Proposition 3 (local uniqueness).} Suppose an encoder uses only bins $k$ and $k+1$ for $t\in[k\Delta,(k+1)\Delta]$. Let $w_k,w_{k+1}\ge0$, $w_k+w_{k+1}=1$, and
\begin{equation}
k\Delta w_k+(k+1)\Delta w_{k+1}=t.
\end{equation}
Then $w_k=1-\alpha$ and $w_{k+1}=\alpha$. Thus local support, mass conservation, and first-moment preservation uniquely recover~\eqref{eq:linear}.

\subsection{Stream-Level Consequence}
Let $\mathcal{E}_\delta$ shift event $i$ from $t_i$ to $t_i+\delta_i$. Triangle inequality and~\eqref{eq:minimax} give
\begin{equation}
\|R_{\rm L}(\mathcal{E}_\delta)-R_{\rm L}(\mathcal{E})\|_1
\le \frac{2}{\Delta}\sum_i |a_i||\delta_i|.
\label{eq:stream}
\end{equation}
If $|\delta_i|\le\epsilon$, normalized drift obeys
\begin{equation}
D(\mathcal{E},\mathcal{E}_\delta)
=\frac{\|R_{\rm L}(\mathcal{E}_\delta)-R_{\rm L}(\mathcal{E})\|_1}{M(\mathcal{E})}
\le \frac{2\epsilon}{\Delta}.
\label{eq:normbound}
\end{equation}
\textbf{Corollary 2 (uniform-jitter expectation).} If the timestamp shifts are independent and $\delta_i\sim\mathcal{U}[-\epsilon,\epsilon]$, then
\begin{equation}
\mathbb{E}\,[D(\mathcal{E},\mathcal{E}_\delta)]\le \frac{\epsilon}{\Delta}.
\label{eq:expected}
\end{equation}
\textit{Proof.} Take expectations in~\eqref{eq:stream} and use $\mathbb{E}|\delta_i|=\epsilon/2$. $\square$ This average-case bound is twice as tight as the deterministic $2\epsilon/\Delta$ bound, although event aggregation can make the observed drift smaller still.

\textbf{Corollary 3 (classifier margin).} Let each pairwise logit difference $g_y-g_c$ have $\ell_1$ Lipschitz constant at most $L_g$. A clean margin $\gamma>2L_g\epsilon M(\mathcal{E})/\Delta$ keeps the predicted class unchanged under the same timestamp budget. We use this only as a sufficient condition; our experiments measure actual flips directly.

\section{Experiments and results}
\subsection{Setup}
We test three event settings. SHD contains 8,156 training and 2,264 test recordings across 20 spoken-digit classes and 700 cochlear spike channels~\cite{cramer2022}. We pool every five channels and use $\Delta=10$ ms. N-MNIST contains 60,000 training and 10,000 test recordings~\cite{orchard2015}. We pool its $34\times34$ sensor to $17\times17$, keep both polarities, and use $\Delta=30$ ms. DVS128 Gesture contains 11 hand and arm gestures recorded from 29 subjects under three lighting conditions~\cite{amir2017}. We use a public presegmented split with 1,077 train and 264 test samples, pool $128\times128$ events to $32\times32$, and use $\Delta=100$ ms.

For every dataset, we compare hard assignment with linear two-bin interpolation. SHD and N-MNIST use small four-layer 2-D CNNs. DVS Gesture keeps time as a separate axis and uses a four-layer 3-D CNN. All networks use batch normalization, GELU, global pooling, and $\log(1+x)$ count compression. AdamW uses learning rate $10^{-3}$ and cosine decay. We report three seeds. We train SHD for 12 epochs, N-MNIST for 8, and DVS Gesture for 25. We did not tune these networks for peak benchmark accuracy.

At test time, we perturb timestamps only. Uniform jitter draws $\delta_i\sim\mathcal{U}[-\epsilon,\epsilon]$ and preserves event count, mark, polarity, and amplitude. We test $\epsilon/\Delta\in\{0.025,0.05,0.1,0.2\}$ on SHD and DVS Gesture, and $\{0.05,0.1,0.2\}$ on N-MNIST. We also test clipped Gaussian jitter on SHD and DVS Gesture.

\begin{table}[t]
\centering
\caption{Clean accuracy and stability at $\epsilon/\Delta=0.1$. Accuracy and flip rate report mean $\pm$ standard deviation over three seeds.}
\label{tab:main}
\resizebox{0.98\linewidth}{!}{%
\begin{tabular}{llccc}
\toprule
Data & Encoding & Clean acc. (\%) & Drift & Flip rate (\%)\\
\midrule
SHD & Hard & $83.60\pm0.96$ & 0.0837 & $3.42\pm0.64$\\
SHD & Linear & $85.22\pm0.53$ & \textbf{0.0372} & $\mathbf{0.88\pm0.28}$\\
N-MNIST & Hard & $99.41\pm0.03$ & 0.0777 & $0.133\pm0.012$\\
N-MNIST & Linear & $99.31\pm0.14$ & \textbf{0.0330} & $\mathbf{0.037\pm0.015}$\\
DVS Gesture & Hard & $95.45\pm0.66$ & 0.0555 & $0.505\pm0.219$\\
DVS Gesture & Linear & $95.20\pm1.16$ & \textbf{0.0214} & $\mathbf{0.000\pm0.000}$\\
\bottomrule
\end{tabular}}
\vspace{-1mm}
\end{table}

\subsection{Timestamp Jitter Across Modalities}
Table~\ref{tab:main} shows the representative budget $\epsilon/\Delta=0.1$. Linear interpolation cuts mean drift by 55.6\% on SHD, 57.6\% on N-MNIST, and 61.4\% on DVS Gesture. It also lowers prediction flips on all three datasets. On DVS Gesture, none of the three linear-interpolation models changes a prediction at any tested uniform budget. Clean accuracy changes little: linear interpolation gains 1.62 points on SHD, loses 0.10 points on N-MNIST, and loses 0.25 points on DVS Gesture.

\begin{figure*}[t]
\centering
\includegraphics[width=0.65\textwidth]{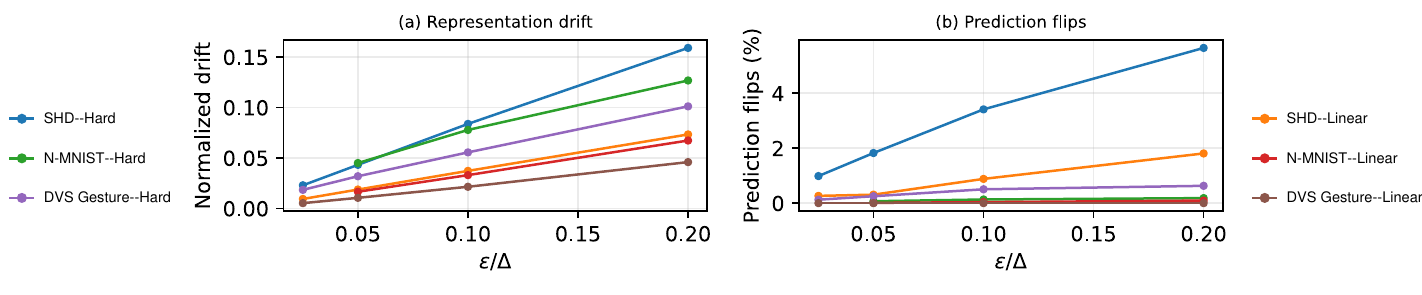}
\vspace{-3mm}
\caption{Timestamp sensitivity across three event settings. 
Linear interpolation lowers (a) normalized representation drift and (b) prediction flips at every tested budget.}
\label{fig:cross}
\vspace{-1.8mm}
\end{figure*}

Figure~\ref{fig:cross} shows the full sweep. Linear interpolation lowers mean drift by 54--59\% on SHD, 47--63\% on N-MNIST, and 55--72\% on DVS Gesture. The median drift shows the same separation. At $0.1\Delta$, DVS median drift falls from 0.0544 to 0.0208, and maximum drift falls from 0.0883 to 0.0437. The separation therefore does not come from a few outliers.

\subsection{Resolution Scaling and Bound Check}
\begin{figure}[t]
\centering
\includegraphics[width=0.4\linewidth]{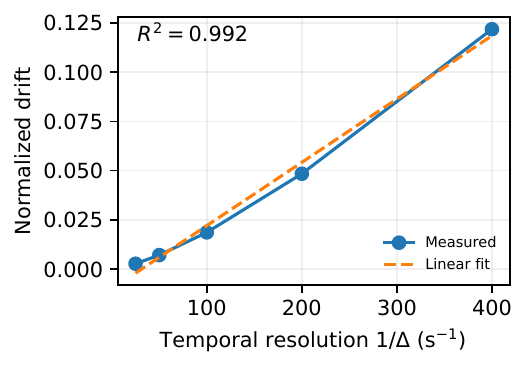}
\vspace{-3mm}
\caption{SHD resolution scaling with fixed $500~\mu$s jitter. Mean drift tracks $1/\Delta$ with $R^2=0.992$.}
\label{fig:resolution}
\vspace{-1.8mm}
\end{figure}

Equation~\eqref{eq:normbound} predicts an inverse dependence on $\Delta$. We test this relation on SHD while fixing absolute jitter at $500~\mu$s. We vary $\Delta\in\{2.5,5,10,20,40\}$ ms. Figure~\ref{fig:resolution} plots measured drift against $1/\Delta$. A linear fit gives $R^2=0.992$. Mean drift rises from 0.00275 at 40 ms to 0.1218 at 2.5 ms. This experiment does not prove the worst-case lower bound, but average sensitivity follows the same inverse-resolution trend.

We also check~\eqref{eq:normbound} directly. Every SHD, N-MNIST, and DVS Gesture sample stays below $2\epsilon/\Delta$ at every tested budget. At $0.2\Delta$, the largest normalized drift reaches 0.111, 0.122, and 0.087 on the three datasets, respectively, while the bound equals 0.4. Across all tested settings, the largest empirical drift uses only 22--31\% of the deterministic bound. We observe zero violations. The mean drift also stays below the sharper uniform-jitter expectation in~\eqref{eq:expected}; at $0.2\Delta$ it reaches 0.073, 0.067, and 0.046 versus an expected bound of 0.2.

\subsection{Perturbation Distribution}
\begin{figure}[t]
\centering
\includegraphics[width=0.86\linewidth]{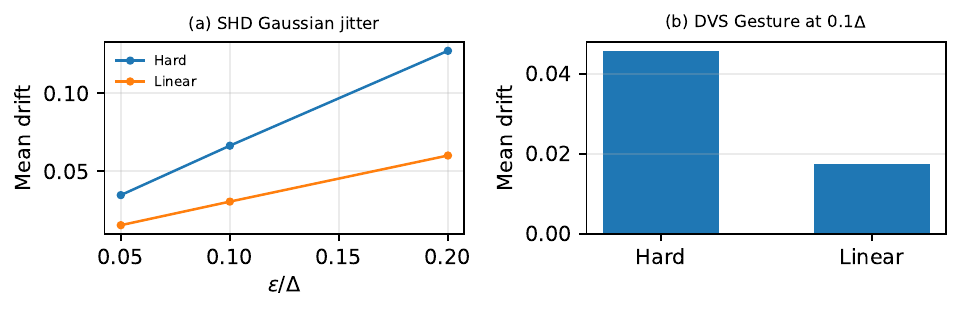}
\vspace{-3mm}
\caption{Clipped Gaussian timestamp jitter. Linear interpolation remains lower on SHD and DVS Gesture.}
\label{fig:gaussian}
\vspace{-5mm}
\end{figure}

Uniform jitter gives the cleanest link to the $\ell_\infty$ budget in~\eqref{eq:normbound}, but timing noise need not follow a uniform law. We repeat the test with zero-mean Gaussian jitter, set its standard deviation to $\epsilon/2$, and clip each sample to $[-\epsilon,\epsilon]$. On SHD, linear interpolation lowers mean drift by 55.8\%, 54.1\%, and 52.8\% at $\epsilon/\Delta=0.05,0.1,$ and $0.2$. At $0.1\Delta$ on DVS Gesture, mean drift falls from 0.0456 to 0.0174, a 61.8\% reduction. The hard models flip 0.25\% of DVS predictions on average, while the linear models again produce zero flips. Figure~\ref{fig:gaussian} summarizes these checks.

\subsection{Moving the Hard-Bin Origin Does Not Fix the Jump}
A simple alternative keeps hard bins and shifts the global bin origin. We sweep 64 evenly spaced phase offsets in $[0,\Delta)$ on SHD. The best phase reduces training-set boundary mass by only 0.74\% relative to the default origin. It changes mean uniform-jitter drift by less than 0.4\% at every budget and gives no consistent flip reduction. Moving the discontinuities therefore does not remove the issue in Proposition~1.

\subsection{Full Perturbation Sweep and Cost}
Table~\ref{tab:sweep} reports every uniform budget. The gap does not depend on one operating point. DVS Gesture gives the widest reduction at small noise: mean drift falls by 71.6\% at $0.025\Delta$. Even at $0.2\Delta$, all three datasets keep a clear separation. DVS also gives an instructive accuracy example. Hard-bin accuracy rises slightly as random jitter changes some decisions, but the model still flips predictions. Linear interpolation keeps every DVS prediction unchanged across all four budgets and all three seeds. Flip rate therefore complements raw accuracy.

The downstream network sees the same tensor shape for both encoders. Linear interpolation changes only event placement. It updates at most two temporal bins per event instead of one. Representation construction therefore adds one weighted update per event while leaving downstream tensor dimensions unchanged.

\begin{table}[t]
\centering
\caption{Uniform timestamp jitter across all tested budgets. Drift gives normalized representation change; flip rates average three seeds.}
\label{tab:sweep}
\resizebox{0.99\linewidth}{!}{%
\begin{tabular}{lcccccc}
\toprule
Data & $\epsilon/\Delta$ & Hard drift & Linear drift & Drop & Hard flip (\%) & Linear flip (\%)\\
\midrule
SHD & .025 & .0229 & .0094 & 59.1\% & .99 & .27\\
SHD & .05  & .0432 & .0187 & 56.7\% & 1.83 & .31\\
SHD & .10  & .0837 & .0372 & 55.6\% & 3.42 & .88\\
SHD & .20  & .1590 & .0733 & 53.9\% & 5.65 & 1.81\\
N-MNIST & .05 & .0450 & .0165 & 63.3\% & .070 & .010\\
N-MNIST & .10 & .0777 & .0330 & 57.6\% & .133 & .037\\
N-MNIST & .20 & .1268 & .0673 & 46.9\% & .183 & .093\\
DVS & .025 & .0184 & .0052 & 71.6\% & .126 & .000\\
DVS & .05  & .0319 & .0105 & 67.1\% & .253 & .000\\
DVS & .10  & .0555 & .0214 & 61.4\% & .505 & .000\\
DVS & .20  & .1011 & .0459 & 54.6\% & .631 & .000\\
\bottomrule
\end{tabular}}
\vspace{-5mm}
\end{table}

\section{Discussion}
\subsection{Interpretation and a Direct Design Rule}
Our analysis targets the temporal encoder, not a specific network. Finer bins can retain more timing detail. Theorem~1 states a narrower fact: if an encoder preserves event mass and exact one-hot anchors spaced by $\Delta$, then its worst-case global timestamp sensitivity cannot fall below $2/\Delta$. Linear interpolation reaches this floor. Hard bins avoid this continuous tradeoff only by becoming discontinuous.

Equation~\eqref{eq:normbound} also gives a direct design rule. If a system expects $|\delta_i|\le\epsilon$ and requires normalized encoder drift below $\tau$, linear interpolation guarantees the target whenever
\begin{equation}
\Delta\ge\frac{2\epsilon}{\tau}.
\label{eq:design}
\end{equation}
For example, $\epsilon=0.5$ ms and $\tau=0.1$ require $\Delta\ge10$ ms. This rule does not select the task-optimal bin width. It states the stability cost that accompanies a chosen temporal resolution.

The lower bound also clarifies how other smooth kernels relate to our result. Higher-order kernels can spread mass across more bins, and learned surfaces can adapt the temporal map to a task~\cite{gehrig2019,cannici2020,lundell2026,liang2025}. If an encoder still preserves nonnegative mass and maps anchors spaced by $\Delta$ to exact one-hot bins, its global $\ell_1$ Lipschitz constant cannot fall below $2/\Delta$. To move below that value, a method must relax at least one condition.

\subsection{Consistency Across Three Event Geometries}
The comparison stays consistent despite very different input geometry. SHD forms a frequency--time tensor, N-MNIST folds polarity and time into image channels, and DVS Gesture keeps polarity, time, height, and width before a 3-D CNN. At $0.1\Delta$, linear interpolation lowers drift by 55.6\%, 57.6\%, and 61.4\%, respectively. Clean accuracy stays within 0.25 percentage points between encoders on N-MNIST and DVS Gesture, while SHD gains 1.62 points. DVS Gesture also adds several-second sequences from different subjects and lighting conditions~\cite{amir2017}; linear interpolation keeps every DVS prediction fixed across four uniform budgets and all three seeds. We treat this as empirical evidence, not a certificate; the deterministic guarantee remains~\eqref{eq:normbound}.

\subsection{Scope and Limitations}
The three datasets cover cochlear spikes, saccadic event vision, and real dynamic gestures. DVS Gesture also adds longer sequences, multiple subjects, and varied lighting. These tests still do not cover every timing-sensitive task. Detection, optical flow, reconstruction, or closed-loop control may show different task-level effects~\cite{kong2025,ye2025,gao2025,kamal2026}. N-MNIST also saturates near 99.4\% clean accuracy, so its absolute flip rates stay small. We therefore treat normalized representation drift as the primary quantity and prediction flips as downstream evidence.

Our deterministic bound controls worst-case $\ell_1$ drift, while random jitter rarely reaches it. Two effects create slack. First, random jitter usually uses less than the full budget $\epsilon$. Second, the proof applies triangle inequality before event aggregation; event changes can partially cancel after they land in the same mark--time tensor. A distribution-specific analysis could predict average drift more tightly.

We also constrain the lower bound to exact, uniformly spaced anchors. Nonuniform grids, learned anchor locations, or encoders that intentionally blur anchor states fall outside $\mathcal{A}_\Delta$. The theorem therefore describes a precise encoder class rather than every possible temporal representation.

The three tasks also span very different difficulty levels. N-MNIST nearly saturates clean accuracy, DVS Gesture reaches about 95\%, and SHD remains much harder. Yet the representation-level ordering stays unchanged. This consistency matters because it separates the encoder effect from a particular classifier accuracy regime. The task metric can saturate or fluctuate, while normalized drift still exposes how the temporal map reacts to the same timestamp budget.

\section{Conclusion}
Hard temporal bins create finite jumps at boundaries. Mass preservation and exact anchors impose a $2/\Delta$ sensitivity floor on continuous encoders, and linear interpolation reaches it. SHD, N-MNIST, and DVS128 Gesture show the same ordering across budgets, modalities, and two noise laws. Linear interpolation cuts representation drift with little clean-accuracy change and sharply lowers prediction flips. The agreement across a cochlear stream, saccadic vision, and real dynamic gestures suggests that the effect comes from temporal encoding rather than any particular dataset. Temporal resolution therefore carries a measurable sensitivity cost and should not enter an event pipeline as a neutral preprocessing choice.


\balance

\clearpage
\begin{thebibliography}{99}
\bibitem{lichtsteiner2008}
P. Lichtsteiner, C. Posch, and T. Delbruck, ``A $128\times128$ 120 dB 15 $\mu$s latency asynchronous temporal contrast vision sensor,'' \emph{IEEE J. Solid-State Circuits}, vol. 43, no. 2, pp. 566--576, 2008.

\bibitem{gallego2022}
G. Gallego, T. Delbruck, G. Orchard, \emph{et al.}, ``Event-based vision: A survey,'' \emph{IEEE Trans. Pattern Anal. Mach. Intell.}, vol. 44, no. 1, pp. 154--180, 2022.

\bibitem{sironi2018}
A. Sironi, M. Brambilla, N. Bourdis, X. Lagorce, and R. Benosman, ``HATS: Histograms of averaged time surfaces for robust event-based object classification,'' in \emph{Proc. IEEE/CVF CVPR}, 2018, pp. 1731--1740.

\bibitem{gehrig2019}
D. Gehrig, A. Loquercio, K. G. Derpanis, and D. Scaramuzza, ``End-to-end learning of representations for asynchronous event-based data,'' in \emph{Proc. IEEE/CVF ICCV}, 2019, pp. 5633--5643.

\bibitem{zhu2019}
A. Z. Zhu, L. Yuan, K. Chaney, and K. Daniilidis, ``Unsupervised event-based learning of optical flow, depth, and egomotion,'' in \emph{Proc. IEEE/CVF CVPR}, 2019, pp. 989--997.

\bibitem{cannici2020}
M. Cannici, M. Ciccone, A. Romanoni, and M. Matteucci, ``A differentiable recurrent surface for asynchronous event-based data,'' in \emph{Proc. ECCV}, 2020.

\bibitem{lundell2026}
F. Lundell, P.-E. Forssen, M. Wadenb\"ack, and A. Lundmark, ``Efficient multi-timescale event representations for feed-forward object detection,'' arXiv:2609.05049, 2026.

\bibitem{lee2022}
W. Lee and H. Myung, ``Adversarial attack for asynchronous event-based data,'' in \emph{Proc. AAAI}, vol. 36, no. 2, pp. 1237--1244, 2022.

\bibitem{yu2026}
Y. Yu, Q. Zhang, S. Ye, \emph{et al.}, ``Time is all it takes: Spike-retiming attacks on event-driven spiking neural networks,'' in \emph{Proc. ICLR}, 2026.

\bibitem{cramer2022}
B. Cramer, Y. Stradmann, J. Schemmel, and F. Zenke, ``The Heidelberg spiking data sets for the systematic evaluation of spiking neural networks,'' \emph{IEEE Trans. Neural Netw. Learn. Syst.}, vol. 33, no. 7, pp. 2744--2757, 2022.

\bibitem{orchard2015}
G. Orchard, A. Jayawant, G. K. Cohen, and N. Thakor, ``Converting static image datasets to spiking neuromorphic datasets using saccades,'' \emph{Frontiers in Neuroscience}, vol. 9, art. 437, 2015.

\bibitem{amir2017}
A. Amir, B. Taba, D. Berg, \emph{et al.}, ``A low power, fully event-based gesture recognition system,'' in \emph{Proc. IEEE/CVF CVPR}, 2017, pp. 7243--7252.


\bibitem{araghi2025}
H. Araghi, J. van Gemert, and N. Tomen, ``Making every event count: Balancing data efficiency and accuracy in event camera subsampling,'' in \emph{Proc. IEEE/CVF CVPR Workshops}, 2025, pp. 5083--5093.

\bibitem{hay2025}
O. Abdul Hay, S. Alansari, M. Alansari, and Y. Zweiri, ``Comparing representations for event camera-based visual object tracking,'' in \emph{Proc. IEEE/CVF ICCV Workshops}, 2025, pp. 4686--4695.

\bibitem{liang2025}
Q. Liang, Q. Li, S. Liu, \emph{et al.}, ``Efficient event camera data pretraining with adaptive prompt fusion,'' in \emph{Proc. IEEE/CVF ICCV}, 2025, pp. 8656--8667.

\bibitem{ye2025}
Y. Ye, H. Shi, K. Yang, \emph{et al.}, ``Towards anytime optical flow estimation with event cameras,'' \emph{Sensors}, vol. 25, no. 10, art. 3158, 2025.

\bibitem{lin2025}
G. Lin, M. Niu, Q. Zhu, \emph{et al.}, ``Adversarial attacks on event-based pedestrian detectors: A physical approach,'' in \emph{Proc. AAAI}, vol. 39, no. 5, pp. 5227--5235, 2025.

\bibitem{gao2025}
Q. Gao, P. Duan, H. Lou, \emph{et al.}, ``Unified reconstruction of static and dynamic scenes from events,'' in \emph{Proc. IEEE/CVF CVPR}, 2025, pp. 27914--27923.


\bibitem{ren2026survey}
H. Ren, Y. Jiang, T. Huang, and X. Wu, ``A systematic survey on event camera representation learning,'' arXiv:2606.23078, 2026.

\bibitem{kong2025}
L. Kong, D. Lu, X. Xu, L. X. Ng, W. T. Ooi, and B. R. Cottereau, ``EventFly: Event camera perception from ground to the sky,'' in \emph{Proc. IEEE/CVF CVPR}, 2025, pp. 1472--1484.

\bibitem{kamal2026}
U. Kamal and S. Mukhopadhyay, ``Memory-augmented representation for efficient event-based visuomotor policy learning with adaptive perception and control,'' in \emph{Proc. IEEE/CVF WACV}, 2026, pp. 2596--2605.

\end{thebibliography}
\end{document}